\documentclass[10pt,twocolumn,letterpaper]{article}

 \usepackage[pagenumbers]{cvpr} 

\usepackage{amsmath,amsfonts,bm}

\def\eqref#1{Eq.~~\ref{#1}}

\def\1{\bm{1}}

\DeclareMathAlphabet{\mathsfit}{\encodingdefault}{\sfdefault}{m}{sl}
\SetMathAlphabet{\mathsfit}{bold}{\encodingdefault}{\sfdefault}{bx}{n}

\newcommand{\E}{\mathbb{E}}

\usepackage{amssymb}

\usepackage{graphicx}%
\usepackage{multirow}%
\usepackage{amsmath,amsfonts}%
\usepackage{amsthm}%
\usepackage{mathrsfs}%
\usepackage[title]{appendix}%
\usepackage{xcolor}%
\usepackage{textcomp}%
\usepackage{manyfoot}%
\usepackage{booktabs}%
\usepackage{algorithm}%
\usepackage{algorithmicx}%
\usepackage{algpseudocode}%
\usepackage{listings}%

\usepackage{times}  
\usepackage{helvet} 
\usepackage{courier}  
\usepackage{subcaption} 
\usepackage{xparse}

\usepackage{xcolor}

\usepackage{url}            
\usepackage{amsfonts}       
\usepackage{nicefrac}       
\usepackage{microtype}      

\usepackage{epstopdf}
\usepackage{rotating,booktabs,multirow}
\usepackage{amsmath,amssymb} 
\usepackage{color}
\usepackage{wrapfig,lipsum}
\usepackage{tabularx}

\usepackage{algorithm}
\usepackage{algpseudocode}
\usepackage{mathtools}
\usepackage{caption}

\usepackage{times}  
\usepackage{helvet}  
\usepackage{courier}  
\usepackage{caption} 

\newcommand{\aka}{\textit{a.k.a. }}

\newcommand{\defeq}{\coloneqq}

\newcommand{\Ea}[1]{\E\left[#1\right]}
\newcommand{\Eb}[2]{\E_{#1}\!\left[#2\right]}

\newcommand{\bI}{\mathbf{I}}

\newcommand{\bzero}{\mathbf{0}}

\newcommand{\bx}{\mathbf{x}}

\newcommand{\bepsilon}{{\boldsymbol{\epsilon}}}
\newcommand{\bmu}{{\boldsymbol{\mu}}}

\newcommand{\bSigma}{{\boldsymbol{\Sigma}}}

\definecolor{cvprblue}{rgb}{0.21,0.49,0.74}
\usepackage[pagebackref,breaklinks,colorlinks,allcolors=cvprblue]{hyperref}

\def\paperID{4} 
\def\confName{CVPR}
\def\confYear{2025}

\title{Adversarially Domain-adaptive Latent Diffusion for\\ Unsupervised Semantic Segmentation}

\author{
	Jongmin Yu\textsuperscript{1,2}
	\quad
        Zhongtian Sun\textsuperscript{2,3}
        \quad
	Chen Bene Chi\textsuperscript{4}
	\quad
	Jinhong Yang\textsuperscript{1,5}
	\quad
        Shan Luo\textsuperscript{6,$\dagger$}
	\\
	\small{$^1$ProjectG.AI \quad $^2$University of Cambridge \quad $^3$University of Kent\quad $^4$Tsinghua University\quad 
    $^5$Inje University\quad 
    $^6$King's College London}
	\\
	\small{\texttt{jy522@projectg.ai}}
	\\
}

\begin{document}

\maketitle

\let\thefootnote\relax\footnotetext{$^{\dagger}$Corresponding author: Shan Luo (shan.luo@kcl.ac.uk)}

\begin{abstract}
Semantic segmentation requires extensive pixel-level annotation, motivating unsupervised domain adaptation (UDA) to transfer knowledge from labelled source domains to unlabelled or weakly labelled target domains. One of the most efficient strategies involves using synthetic datasets generated within controlled virtual environments, such as video games or traffic simulators, which can automatically generate pixel-level annotations. However, even when such datasets are available, learning a well-generalised representation that captures both domains remains challenging, owing to probabilistic and geometric discrepancies between the virtual world and real-world imagery. This work introduces a semantic segmentation method based on latent diffusion models, termed Inter-Coder Connected Latent Diffusion (ICCLD), alongside an unsupervised domain adaptation approach. The model employs an inter-coder connection to enhance contextual understanding and preserve fine details, while adversarial learning aligns latent feature distributions across domains during the latent diffusion process. Experiments on GTA5, Synthia, and Cityscapes demonstrate that ICCLD outperforms state-of-the-art UDA methods, achieving mIoU scores of 74.4 (GTA5$\rightarrow$Cityscapes) and 67.2 (Synthia$\rightarrow$Cityscapes).
\end{abstract}    
\section{Introduction}
\label{sec:intro}
Semantic segmentation, which involves assigning a semantic label to each pixel in an image, is crucial for applications such as autonomous driving. Despite significant advances in deep learning on standard benchmarks \cite{he2022masked, he2016deep,ding2023mose,ding2023mevis}, many state-of-the-art (SOTA) models still struggle in real-world environments due to domain variations such as differences in lighting, geographic location, and weather conditions. Although increasing the diversity of training data through manual annotation can improve performance, the high cost, for example, approximately 90 minutes per image in Cityscapes \cite{cordts2016Cityscapes}, renders this approach impractical. Synthetic datasets such as Synthia \cite{ros2016synthia} and GTA5 \cite{richter2016playing} offer a promising alternative. As these datasets are generated within controlled virtual environments, such as video games \cite{richter2016playing}, pixel-level annotations can be automatically generated. However, the substantial visual disparity between simulated and real domains can significantly degrade the performance of models trained solely on synthetic data.

Various domain adaptation techniques have been proposed to address this issue by transferring knowledge from a labelled source domain to unlabelled or weakly labelled target domains. These techniques are typically categorised as supervised \cite{li2019bidirectional, wang2021loveda}, semi-supervised \cite{french2019semi, hoyer2021three, lai2021semi, souly2017semi, dai2015boxsup, song2019box, zou2020pseudoseg}, self-supervised \cite{vayyat2022cluda}, or unsupervised \cite{hoffman2016fcns, tsai2018learning, zou2018unsupervised}, depending on the availability of target domain labels. Recent approaches have further enhanced model generalisation by incorporating advanced network architectures, such as transformers \cite{hoyer2021daformer} and attention mechanisms \cite{yu2023multi}, as well as learning strategies like teacher and student frameworks \cite{hoyer2021daformer, vayyat2022cluda}.

This paper addresses the challenge of unsupervised domain adaptation (UDA) in semantic segmentation. UDA aims to adapt a model trained on a labelled source domain to perform effectively on an unlabelled target domain, which is a scenario that presents significant challenges due to the absence of direct and explicit supervision in the target domain \cite{ren2020learning}. Prior research has explored a variety of network architectures and objective functions to mitigate these challenges \cite{hoffman2016fcns, tsai2018learning, zou2018unsupervised, hoyer2021daformer, vayyat2022cluda}.

To address this challenge, we propose a novel semantic segmentation framework that integrates a diffusion model with adversarial learning. Specifically, we present Inter-Coder Connected Latent Diffusion (ICCLD), which incorporates long skip connections to bridge information between each encoder and decoder pair, effectively combining low- and high-level features with the latent representation produced by the diffusion process \cite{ho2020denoising, yu2023adversarial}.

Additionally, in the context of UDA, ICCLD employs a student and teacher learning framework \cite{vayyat2022cluda}, whereby the model iteratively refines its predictions on the target domain. Adversarial learning is integrated during the denoising phase, where a denoising UNet is trained to align feature distributions between the source and target domains by confusing a domain discriminator.

In our experiments, ICCLD achieves SOTA performance on two challenging unsupervised domain adaptation benchmarks. Specifically, it attains a mean Intersection-over-Union (mIoU) of 74.4 for the GTA5→Cityscapes setting and 67.2 for the Synthia→Cityscapes setting, surpassing existing state-of-the-art UDA methods. These results underscore the superior domain adaptation capabilities of the proposed approach and its potential to narrow the performance gap between source and target domains in semantic segmentation tasks.

The primary contributions of this work are as follows:

\begin{itemize}
    \item \textbf{Latent diffusion-based semantic segmentation model (ICCLD):} We introduce ICCLD, which leverages long skip connections between encoder and decoder modules, combined with a conditioning mechanism, to enable the learning of more accurate semantic representations.
    \item \textbf{Improved domain adaptation via adversarial learning:} Our framework incorporates adversarial learning into the denoising process of the latent diffusion model, aligning latent feature distributions between the source and target domains.
    \item \textbf{Comprehensive empirical validation:} Extensive ablation studies and comparisons with existing methods demonstrate that ICCLD achieves superior performance on unsupervised domain adaptation benchmarks, with mIoU scores of 74.4 (GTA5$\rightarrow$Cityscapes) and 67.2 (Synthia$\rightarrow$Cityscapes).
\end{itemize}

\section{Related works}
\label{sec:rw}
\subsection{Semantic Segmentation}
Semantic segmentation assigns a label to every pixel in an image and is essential for applications such as autonomous driving \cite{xie2021segformer}, medical imaging \cite{ronneberger2015u, chen2021transunet}, and scene understanding \cite{strudel2021segmenter}. Since the advent of deep learning, convolutional neural networks (CNNs) \cite{lecun1989handwritten} have driven significant advancements in this field. Shelhamer et al. \cite{shelhamer2016fully} present a semantic segmentation model composed solely of convolutional layers and demonstrate that CNNs provide highly effective representation learning for this task. Noh et al. \cite{noh2015learning} and Visin et al. \cite{visin2016reseg} also propose CNN-based segmentation methods. However, the locality of the learnt representations in CNNs is often regarded as a limitation in semantic segmentation.

To address this issue, transformer-based methods have been introduced. Initially developed for natural language processing, transformers are now pivotal in computer vision tasks. Models such as Segmenter \cite{strudel2021segmenter}, SegFormer \cite{xie2021segformer}, and TransUNet \cite{chen2021transunet} leverage transformer architectures to capture long-range dependencies, thereby enhancing segmentation accuracy. While these methods benefit from improved global context modelling, they typically demand greater computational resources and memory than traditional CNN-based approaches.

Recently, diffusion models have shown promise in generative modelling and are gaining attention in semantic segmentation \cite{song2020score, song2020denoising}. The Denoising Diffusion Implicit Model (DDIM) \cite{song2020denoising} demonstrates the potential for high-fidelity image generation, which can be adapted for semantic segmentation tasks. Recent works \cite{yu2023adversarial, baranchuk2021label, tan2022semantic, wu2023diffumask, amit2021segdiff} have begun to explore diffusion processes for generating multi-scale latent features and improving segmentation outcomes. These approaches typically assume access to large-scale, well-organised training datasets. In this work, we extend this line of research by introducing a semantic segmentation method based on diffusion models within an unsupervised domain adaptation framework.

\subsection{Unsupervised Domain Adaptation}
Unsupervised Domain Adaptation (UDA) focuses on transferring knowledge from a labelled source domain to an unlabelled target domain \cite{hoyer2021daformer}. In the context of semantic segmentation, UDA techniques have progressed through strategies such as data augmentation \cite{wang2022cross, hoffman2018cycada}, feature alignment \cite{saito2020adversarial, yu2023adversarial, peng2024unsupervised}, and self-supervised learning \cite{hoyer2021daformer, yu2023multi}. Synthetic datasets such as GTA5 \cite{richter2016playing} and Synthia \cite{ros2016synthia} are commonly used as the labelled source domain, with the goal of transferring learned representations to real-world datasets like Cityscapes. Although synthetic datasets are significantly easier to generate than manually annotated real-world images, they exhibit notable differences in texture, object appearance, and contextual structure. This domain discrepancy underscores the need for methods capable of learning highly generalised representations from synthetic data.

It is well established that the generalisation ability of segmentation models often does not transfer effectively across domains, leading to suboptimal cross-domain performance. To address this issue, a range of adaptation methods have been proposed, including adversarial training at the input image level \cite{hoffman2018cycada}, feature level \cite{hoffman2016fcns, hoffman2018cycada}, and output level \cite{tsai2018learning}. For example, Hoffman et al. \cite{hoffman2018cycada} reduce the domain gap by translating source images into the style of the target domain using a cycle-consistency loss, followed by adversarial alignment of features. Similarly, Saito et al. \cite{saito2020adversarial} propose a critic network to identify non-discriminative samples near decision boundaries, thereby encouraging the generator to learn more discriminative features. Curriculum adaptation methods \cite{zhang2017curriculum} further constrain the predicted label distributions to promote alignment between source and target domains.

To enhance segmentation performance in UDA scenarios, we propose a semantic segmentation approach based on latent diffusion models, integrated with adversarial learning. Our method exploits the capacity of diffusion models to learn generalised representations from synthetic images. Adversarial learning is employed to reduce the domain gap in the latent feature space, thereby facilitating a more effective transfer of knowledge from the source to the target domain.

\begin{figure*}[t]
\begin{center}
\includegraphics[width=\linewidth]{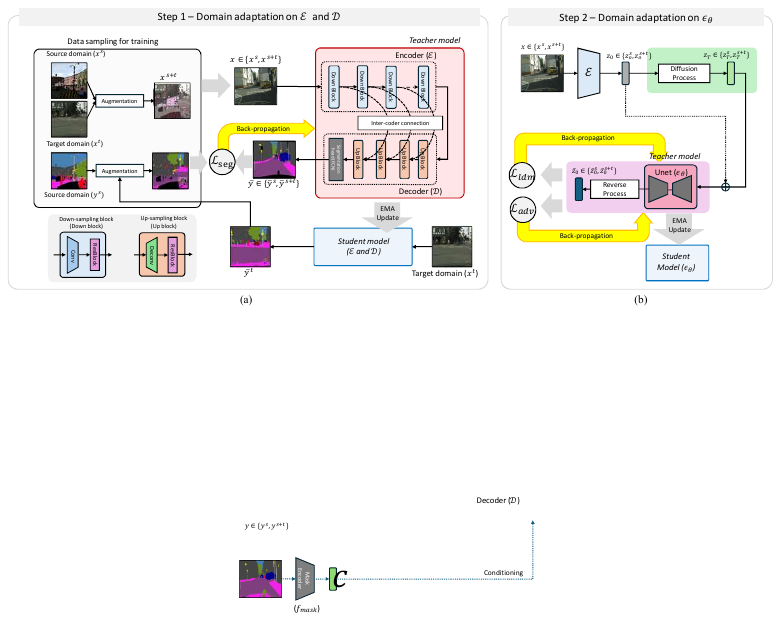}  
\vspace{-2ex}
\end{center}
    \caption{Illustration of the workflow for the two-step domain adaptation process using the proposed Inter-Coder Connected Latent Diffusion (ICCLD) framework. In the first stage, domain adaptation is performed on the encoder $\mathcal{E}$ and decoder $\mathcal{D}$ of ICCLD using a segmentation-based approach. ClassMix method \cite{olsson2021classmix} is used as a data augmentation method by generating mixed images $x^{s+t}$  and labels $y^{s+t}$. In the second stage, adversarial learning is applied to the denoising UNet $\epsilon_{\theta}$ for further domain alignment. During both stages, the loss functions are primarily computed using the teacher model, while the student model is updated via an exponential moving average (EMA) of the teacher model parameters.}
\label{fig:full_model}
\vspace{-2ex}
\end{figure*}

\section{The proposed method}
\label{sec:method}
\subsection{Preliminaries}
The goal of diffusion models \cite{ho2020denoising} is to find the parameterised data distribution $p_\theta(\bx_0)$ using a given data $\bx_0 \sim q(\bx_0)$. To do this, when given data point $\bx_0$, the training of diffusion models conducts a \emph{forward process} (\aka diffusion process) $q(\bx_{ \text{t}} | \bx_{ \text{t}-1})$, which adds Gaussian noise to the data, and a \emph{reverse process} (\aka de-noising process) $p_\theta(\bx_{ \text{t}-1}|\bx_ \text{t})$, which denoises the given noised data by subtracting predicted noise.

The forward process $q(\bx_{ \text{t}} | \bx_{ \text{t}-1})$ is a task to add Gaussian noise to data at a certain time step $t \leq  \text{T}$. The noise vector is generated by a Gaussian distribution $\mathcal{N}(\cdot)$ with scheduled variance: $\beta_1, \dotsc, \beta_ \text{T}$. The entire forward process to generate completely noised sample $\bx_{ \text{T}}$ is represented by 
\begin{equation}
\begin{aligned}
& q(\bx_{1: \text{T}} | \bx_0) \defeq \prod_{ \text{t}=1}^ \text{T} q(\bx_ \text{t} | \bx_{ \text{t}-1} ), \qquad \\ &
q(\bx_ \text{t}|\bx_{ \text{t}-1}) \defeq \mathcal{N}(\bx_ \text{t};\sqrt{1-\beta_ \text{t}}\bx_{ \text{t}-1},\beta_\text{t} \bI).
\end{aligned}
\label{eq:forwardprocess}
\end{equation}

The reverse process $p_\theta(\bx_{ \text{t}-1}|\bx_ \text{t})$ can be considered a de-noising task. For each time step $ \text{t}$, a diffusion model predicts a noise and subtracts it from the noised data. This task is represented by Markov Chain so that it can be represented by parametric conditional distribution, as follows:
\begin{equation}
\begin{aligned}
  & p_\theta(\bx_{0: \text{T}}) \defeq p(\bx_ \text{T})\prod_{ \text{t}=1}^ \text{T}  p_\theta(\bx_{ \text{t}-1}|\bx_ \text{t}), \qquad 
  \\ &p_\theta(\bx_{ \text{t}-1}|\bx_ \text{t}) \defeq \mathcal{N}(\bx_{ \text{t}-1}; \bmu_\theta(\bx_ \text{t},  \text{t}), \bSigma_\theta(\bx_ \text{t},  \text{t})).
\end{aligned}\
\label{eq:reverseprocess}
\end{equation}

The learning of the diffusion model is to find the suitable parametric distribution $p_\theta(\bx_{0})$ representing a given data. This distribution is central to reconstructing the original data from its noised state. However, directly optimising the negative log-likelihood of $p_\theta(\bx_{0})$ poses computational challenges due to its complexity. Consequently, optimisation is conducted via a variational lower bound, which is more computationally feasible and is formulated as follows:
\begin{equation}
\begin{aligned}
\mathcal{L}_{dm}  & \defeq  \Ea{-\log p_\theta(\bx_0)} \leq \Eb{q}{ - \log \frac{p_\theta(\bx_{0: \text{T}})}{q(\bx_{1: \text{T}} | \bx_0)}}
 \\& = \mathbb{E}_q\bigg[ -\log p(\bx_ \text{T}) - \sum_{ \text{t} \geq 1} \log \frac{p_\theta(\bx_{ \text{t}-1} | \bx_ \text{t})}{q(\bx_ \text{t}|\bx_{ \text{t}-1})} \bigg].
\end{aligned}
\label{eq:dm_loss}
\end{equation}

In DDPM \cite{ho2020denoising}, it was shown that reparameterisation allows the forward and reverse processes to be expressed in terms of noise prediction. Consequently, it can be replaced by minimising prediction error for a Gaussian noise and noise prediction obtained by a neural network. As a result, \eqref{eq:dm_loss} is further simplified to the $l2$-distance between the generated Gaussian noise $\bepsilon \sim \mathcal{N}(\bzero, \bI)$ and predicted noise $\bepsilon_\theta$ at a certain time step. Using the notations, $\alpha_ \text{t} \defeq 1-\beta_ \text{t}$ and $\bar\alpha_t \defeq \prod_{s=1}^ \text{t} \alpha_s$, the simplified loss is represented by 
\begin{equation}
\begin{aligned}
  \mathcal{L}_{ldm} = \Eb{t, \bx_0, \bepsilon}{ \left\| \bepsilon - \bepsilon_\theta(\sqrt{\bar\alpha_t} \bx_0 + \sqrt{1-\bar\alpha_t}\bepsilon, t) \right\|^2}.
\end{aligned}
\label{eq:ddpm_loss}
\end{equation}

\subsection{Methodology overview}
Figure~\ref{fig:full_model} illustrates our proposed approach for unsupervised domain adaptation using the Inter-Coder Connected Latent Diffusion (ICCLD) framework. The training process is conducted in two stages: (1) domain adaptation via segmentation, and (2) latent distribution alignment using a latent diffusion model.

In the first stage, domain adaptation via segmentation is applied to derive domain-adapted encoder $\mathcal{E}$ and decoder $\mathcal{D}$. These components are trained to minimise pixel-wise classification error using a segmentation-based approach. To facilitate this, we generate synthetic images and corresponding labels by combining annotated source domain data with unlabelled target domain data. The synthetic data enable the model to learn supervisory signals applicable to the target domain.

In the second stage, latent distribution alignment is performed by training the denoising network $\epsilon_{\theta}$. The network is primarily optimised using the noise prediction loss defined in Equation~\ref{eq:ddpm_loss}. To further reduce the domain discrepancy in the latent space, we introduce an additional loss term based on adversarial learning. This loss encourages the predicted noise distributions for the source domain to align with those of synthetic target-domain samples, thereby mitigating probabilistic divergence.

Throughout both training stages, gradients from the segmentation loss and the adversarial learning loss are back-propagated only through the teacher model. The student model is updated via the Exponential Moving Average (EMA) of the teacher model’s parameters, promoting stable training and improved generalisation \cite{hoyer2021daformer}. During inference, only the student model is used for prediction.

\subsection{Architectural Details of ICCLD}
Latent Diffusion Models (LDMs) \cite{rombach2022high} have demonstrated strong performance in conditional image translation tasks \cite{baranchuk2021label, tan2022semantic, wu2023diffumask, amit2021segdiff}. An LDM typically consists of three core modules: (1) an encoder, (2) a denoising UNet, and (3) a decoder. The encoder embeds the input into a latent feature space while the decoder reconstructs outputs in the target domain. The denoising UNet iteratively refines the latent representation, transforming it into a well-known distribution, such as the Gaussian distribution, through a sequence of forward and reverse diffusion steps. This architecture enables LDMs to significantly reduce computational overhead compared to conventional diffusion models operating directly in high-dimensional image space. Moreover, LDMs facilitate flexible conditioning via self-attention mechanisms embedded in the denoising network $\epsilon_{\theta}$.

Inspired by the LDM structure, our proposed Inter-Coder Connected Latent Diffusion (ICCLD) model also comprises three main components: (1) an encoder $\mathcal{E}$, (2) a denoising UNet $\epsilon_{\theta}$, and (3) a decoder $\mathcal{D}$. The encoder $\mathcal{E}$ extracts latent features from input images, and the decoder $\mathcal{D}$ upsamples these features to produce the final segmentation mask. The denoising network $\epsilon_{\theta}$ performs the diffusion and denoising operations to model the latent feature distribution. During inference, $\mathcal{D}$ is used to generate a domain-adapted latent representation conditioned on the input image.

A key distinction between ICCLD and standard LDMs lies in the explicit connection established between the encoder and the decoder. While conventional LDMs include skip connections within the internal structure of $\epsilon_{\theta}$, these connections are restricted to the UNet itself and do not directly link the encoder and decoder modules. Consequently, although semantic information can be retained through conditioning, the decoder’s upsampling process may lack access to the fine-grained spatial detail required to delineate object boundaries accurately.

To address this limitation, ICCLD introduces an inter-coder connection that directly feeds abstracted features from the encoder to the decoder. This connection allows rich spatial hierarchies and fine details to propagate through the network, improving the model’s ability to capture textural and boundary-specific information. Prior studies \cite{zhou2019unet++, wang2022uctransnet} have shown that long skip connections are effective at preserving semantic and contextual features, which in turn enhance segmentation accuracy. Our approach leverages this insight to explicitly bridge encoder and decoder representations, thereby enhancing segmentation performance.

We empirically validate the effectiveness of the inter-coder connection through ablation studies presented in Section~\ref{effective_intercoder} and demonstrate its contribution to improved boundary delineation and overall segmentation quality.
\begin{figure}
\begin{center}
\includegraphics[width=\columnwidth]{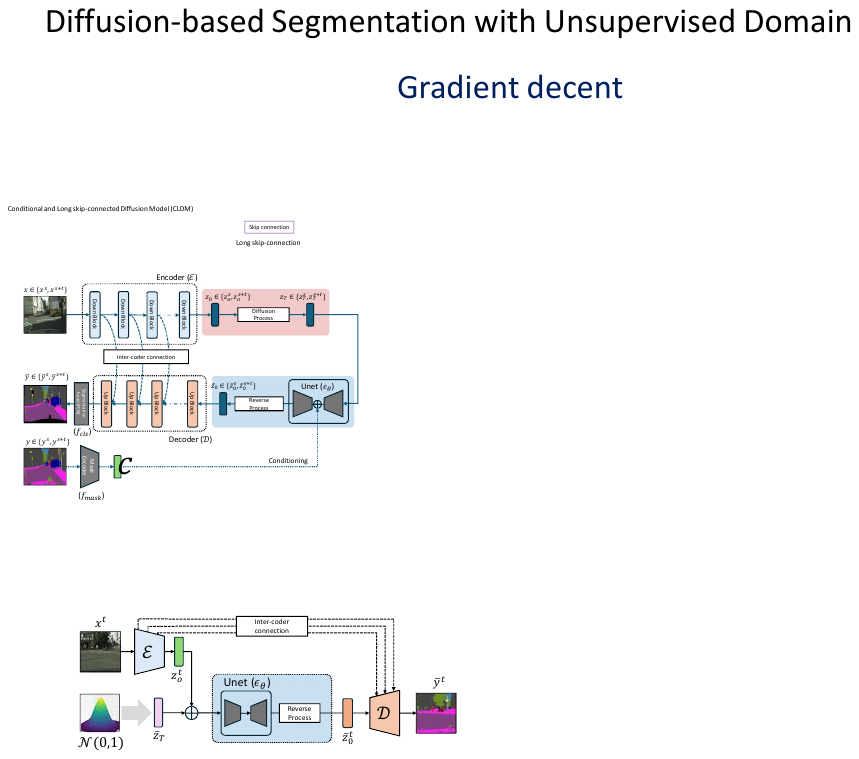}  
\vspace{-2ex}
\end{center}
\caption{Architectural details of ICCLD. For prediction a segmentation mask $\bar{y}^{t}$, ICCLD samples a noise vector $\bar{z}_{T}$ from Gaussian distribution $\mathcal{N}(0,1)$, and extracts a latent vector $z^{t}_{0}$ from a given image $x^{t}$. $\bar{z}_{T}$ and $z^{t}_{0}$ are concatenated and applied to the de-noising UNet $\epsilon_{\theta}$ for generating a latent feature vector $\bar{z}^{t}_{0}$. The de-noising process is repeated for $T$ times to obtain $\bar{z}^{t}_{0}$. The decoder $\mathcal{D}$ takes $\bar{z}^{t}_{0}$ as an input and predicts $\bar{y}^{t}$.}
\label{fig:archi_details}
\vspace{-2ex}
\end{figure}

\subsection{Training and Domain Adaptation with ICCLD}
\label{objfunction}
Unsupervised domain adaptation (UDA) for semantic segmentation using ICCLD consists of two distinct phases. The first phase involves domain adaptation for the encoder $\mathcal{E}$ and decoder $\mathcal{D}$ using a segmentation-based approach. In the second phase, the denoising network $\epsilon_{\theta}$ is trained to minimise the domain gap between the source and target domains by leveraging latent diffusion in conjunction with adversarial learning.

In this paper, we omit the training procedure for $\mathcal{E}$ and $\mathcal{D}$ based on conventional segmentation methods using a labelled source domain. The details of each phase are provided below.

\noindent
\textbf{Step 1-Domain adaptation on $\mathcal{E}$ and $\mathcal{D}$:} After finishing the training of $\mathcal{E}$ and $\mathcal{D}$ using the labelled source domain, we copy the trained parameters to make a student model and use the original model as a teacher model. After that, our framework takes as input a source image \(x^{s}\) and a target image \( x^{t} \), sampled from a source dataset and a target dataset, respectively. A mixed image \( x^{s+t} \) is then generated using the ClassMix method \cite{olsson2021classmix}, where source pixels from randomly selected classes are overlaid on the target image with pixel-level augmentations. This mixing process is equivalently applied to the corresponding labels. Since the target domain lacks annotations during training, pseudo labels \( y^{s+t} \) for \( x^{s+t} \) are generated by combining predicted labels $\bar{y}^{t}$ obtained from the student model (see Figure \ref{fig:full_model}) and prepared target label $y^{s}$. The teacher learns information about the target domain by using $x^{s+t}$ and $\bar{y}^{t}$ and delivers it to the student model. The loss function using $x^{s}$, $x^{s+t}$, $y^{s}$, and  $y^{s+t}$ is formulated based on the cross-entropy function, which is defined as follows:
\begin{equation}
\label{eq:class}
\mathcal{L}_{seg}(\bar{y}^{*},y^{*}) = -\E \left[ y^{*} \log(\bar{y}^{*}) \right],
\end{equation}
where $\bar{y}^{*}$ can be defined by predicted segmentation results $\bar{y}^{s}$ and $\bar{y}^{s+t}$, and $y^{*}$ can be defined by image labels $y^{s}$ and $y^{s+t}$. Figure \ref{fig:full_model}(a) provides a graphical explanation of the first step of DA for $\mathcal{E}$ and $\mathcal{D}$.

\noindent
2\textbf{Step 2-Adversarial domain adaptation on $\epsilon_{\theta}$:} The second phase is applied to improve domain adaptation performance by explicitly aligning latent feature distributions when we train the de-noising Unet $\epsilon^{\theta}$. $\mathcal{E}$ are frozen during the second step. The diffusion loss function in the DDPM \cite{ho2020denoising} (Eq. \ref{eq:ddpm_loss}) applies the diffusion process to the latent feature space with some modification for some conditioning. The loss function is used to optimise  $\epsilon^{\theta}$ only by minimising given noise $\epsilon$ and the predicted noise by the Unet $\bar\epsilon$. 

Since semantic segmentation is not just generating arbitrary segmentation masks. ICCLD should generate a suitable segmentation mask corresponding with the given image so that conditioning using the latent features is essential. As shown in Figure \ref{fig:full_model}(b), the clear latent feature $z_{0}$, extracted by $\mathcal{E}$, is applied as a conditional factor during diffusion. The loss function for the $t$-step diffusion and de-noising processes is formulated as follows: 
\begin{equation}
\begin{aligned}
  \mathcal{L}_{ldm}(t,z^{*},z^{*}_{0}, \epsilon) ={ \left\| \epsilon - \epsilon_\theta(\sqrt{\bar\alpha_t} z^{*} + \sqrt{1-\bar\alpha_t}\epsilon, t|z^{*}_{0}) \right\|^2},
\end{aligned}
\label{eq:addm_loss}
\end{equation}
where $z^{*}$ can be defined by the latent features for the source image $z^{s}$, mixed image $z^{s+t}$, and target image $z^{t}$. $z^{*}_{0}$ indicates clear latent features that have no noise, and are extracted from source image $z^{s}_{0}$, target image $z^{s+t}_{0}$, and mixed image $z^{s+t}_{0}$.  The scheduled variance defines $\alpha$. Please refer to the DDPM \cite{ho2020denoising} for further details.

An adversarial loss is applied during every diffusion and de-noising process to align the latent features from source, mixed, and target domains. Adversarial learning is formulated to distinguish the predicted noise for each domain. Since adversarial learning takes three classes of inputs considering their domains (source, target, and mixed domains), we formulate the adversarial learning loss using KL-divergence. The adversarial loss is formulated as follows:
\begin{equation}
\begin{aligned}
& \mathcal{L}_{adv}(z^{*},t) = \mathbb{E}\left[ o^{*}\log\left(f_{dis}(\epsilon_\theta(\sqrt{\bar\alpha_t} z^{*} + \sqrt{1-\bar\alpha_t}\epsilon, t) )\right)\right] \\&+ \mathbb{E}\left[D_{KL}\left(f_{dis}(\epsilon_\theta(\sqrt{\bar\alpha_t} z^{*} + \sqrt{1-\bar\alpha_t}\epsilon, t) ) | U(\frac{1}{3})\right)\right],
\end{aligned}
\label{eq:addm_loss}
\end{equation}
where $f_{dis}$ defines a neural network that plays the role of discriminator in computing the adversarial loss, and it outputs the results of the softmax function. $z^{*}$ can be defined by the latent features for the source image $z^{s}$, mixed image $z^{s+t}$, and target image $z^{t}$. $o{*}$ defines pseudo-one-hot encoded vectors for indicating the domain of latent features used for predicting noise. For example, when $z^{s}$ is given, $o^{*}$ is represented as $o^{s}$ and set by $[1,0,0]$. $\mathcal{U}(\frac{1}{3})$ represents a 3D vector to define a uniform distribution defined by 1/3 of each elements. Figure \ref{fig:full_model}(b) presents an illustration to describe the second step of domain adaptation on $\epsilon_{\theta}$.

\noindent
\textbf{Inference:} Once both training phases are completed, inference on the target domain is performed exclusively using the student model; the teacher model is not utilised during this stage. The inference procedure is as follows: a noise vector is first sampled from a Gaussian distribution. This is then concatenated with the latent feature extracted from the input image via the encoder $\mathcal{E}$. The denoising process is subsequently applied for a predefined number of steps using the denoising network $\epsilon_{\theta}$. Finally, the resulting denoised latent feature is passed through the decoder $\mathcal{D}$ to produce the final segmentation mask.

\begin{figure}
\begin{center}
\includegraphics[width=\columnwidth]{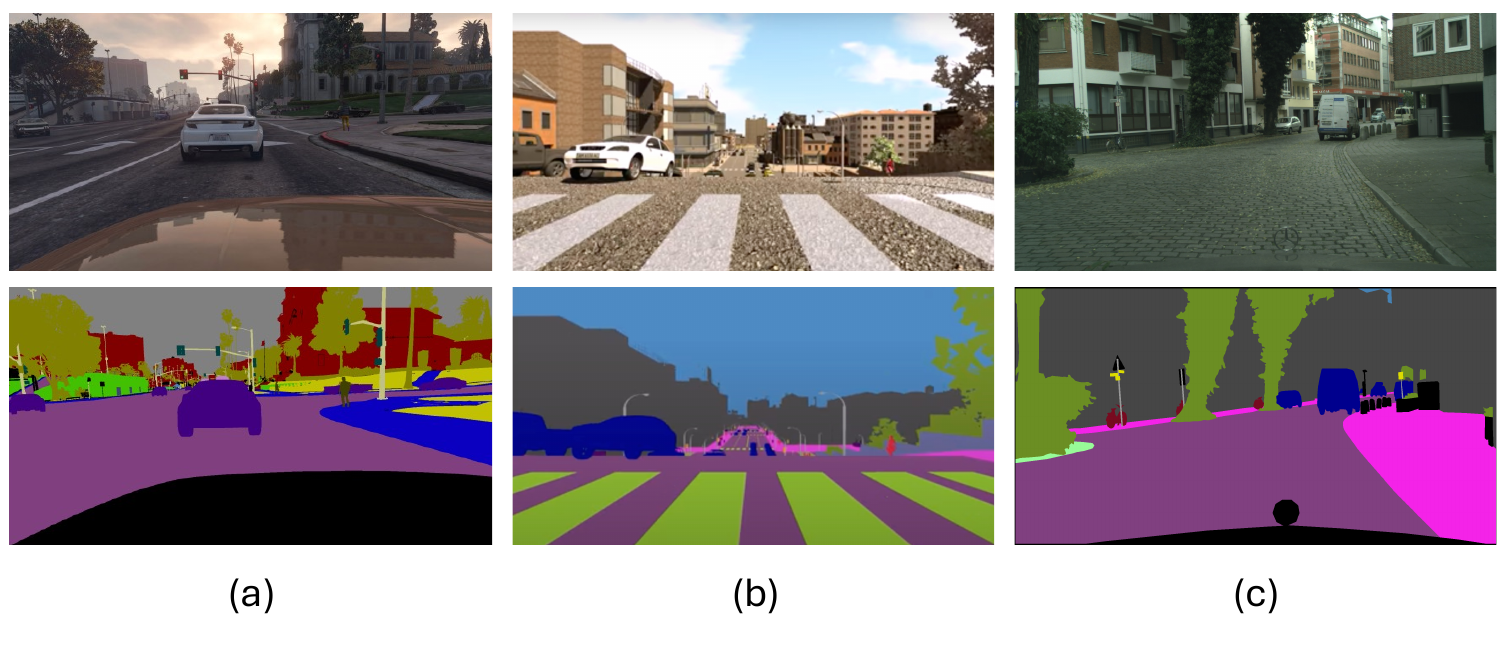}      
\end{center}
\vspace{-2ex}
\caption{Example images and labels of the (a) GTA-5 \cite{richter2016playing}, (b) SYNTHIA \cite{ros2016synthia}, and (c)Cityscapes datasets \cite{cordts2016Cityscapes}.}
\label{fig:examples}
\vspace{-2ex}
\end{figure}

\begin{figure*}
\begin{center}
\includegraphics[width=\linewidth]{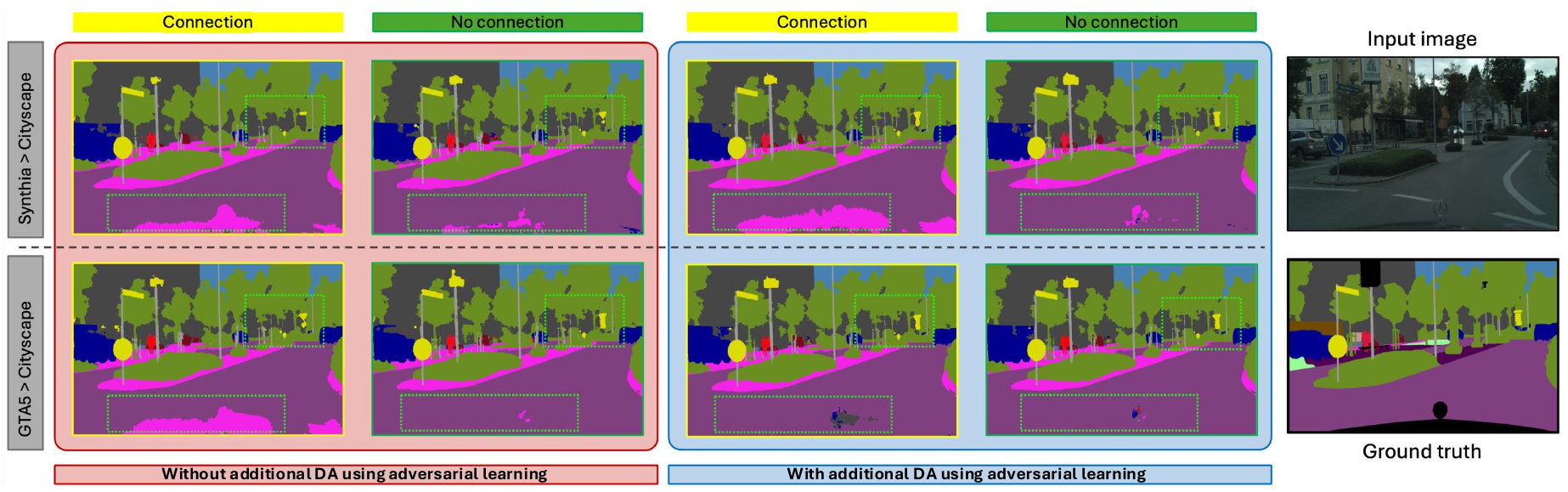}    
\end{center}
\caption{Qualitative comparison of the UDA performance of ICCLD according to the inter-coder connection and the usage of the extra DA process using adversarial learning. The quantitative results highlighted by green-coloured dotted boxes show that the inter-coder connections and the additional DA process improve the quality of segmentation results by reducing false-positive results.}
\label{fig:quant_results_ablation}
\end{figure*}

\section{Experimental settings}
\label{sec:exp}
\noindent
\textbf{Dataset and experimental protocol:} We choose \textbf{GTA5} \cite{richter2016playing}, \textbf{SYNTHIA} \cite{ros2016synthia}, \textbf{Cityscapes} \cite{cordts2016Cityscapes} datasets for our experiments. Figure \ref{fig:examples} shows the example snapshots of images and labels of those three datasets. Those datasets are publicly available. Detailed information on those datasets is as follows.

\noindent
\textbf{GTA-5}\footnote{\url{https://download.visinf.tu-darmstadt.de/data/from_games}} has 24,966 synthetic images extracted from a photo-realistic open-world game called Grand Theft Auto, along with semantic segmentation maps. The image resolution is
1914x1052.

\noindent
\textbf{Synthia}\footnote{\url{https://synthia-dataset.net/}} is a synthetic dataset comprising 9400 photo-realistic frames with a resolution of 1280 × 960, rendered from a virtual city with pixel-level annotations for 13 classes.

\noindent
\textbf{Cityscapes}\footnote{\url{https://www.cityscapes-dataset.com/}} is a large-scale database that focuses on semantic understanding of urban street scenes. The dataset has semantically annotated 2975 training images and 500 validation images with a resolution of 2048x1024.

We evaluate the UDA performances for GTA5-to-Cityscapes (GTA5$\rightarrow$Cityscapes) and Synthia-to-Cityscapes (Synthia$\rightarrow$Cityscapes). The experiment protocol is referred from the CLUDA \cite{vayyat2022cluda} and Toldo \etal \cite{toldo2021unsupervised}.

\noindent
\textbf{Implementation details:} We employ the Latent Diffusion Model (LDM) \cite{rombach2022high} as the backbone.To apply the intercoder connection, the kernel dimensionalities of convolutional layers on the decoder have been modified. The architectural setting of the discriminator $f_{dis}$ is equivalent with the encoding part of the denoising UNet $\epsilon_{\theta}$, but puts an additional neural network for three class classifications. During inference, we use DDIM scheduler \cite{song2020denoising} and set 50 iterations for the denoising process. The Adam optimiser is used for optimisation. The initial learning rate is 6 $\times$ 10$-$5, and the learning rate is decayed every five epochs by multiplying 0.99 by the learning rate. The batch size is 2. Fifty epochs are set for training. For updating the student model, we keep the value of the EMA weight update parameter $\alpha$ = 0.999; for learning-rate optimisation, we follow polynomial-learning rate reduction.

\begin{table}[t]
\resizebox{\columnwidth}{!}{%
    \centering
    \begin{tabular}{@{}l|c|c@{}}
        \toprule
        DA setting & GTA5 $\rightarrow$ Cityscapes & Synthia $\rightarrow$ Cityscapes \\ \midrule
        \midrule
        \multicolumn{3}{c}{Without the Inter-coder connection}\\
        \midrule
        Step 1 (DA on $\mathcal{E}$ and $\mathcal{D}$) &  58.3  &    42.7   \\
        \midrule
        Step 1 \& 2 (DA on $\mathcal{E}$, $\mathcal{D}$, and $\epsilon_{\theta}$) &  69.3  &   58.5     \\
        \midrule
        \multicolumn{3}{c}{With the Inter-coder connection}\\
        \midrule
        Step 1 &  68.4   &  60.7     \\
        \midrule
        Step 1 \& 2 &  \textbf{74.4}   &   \textbf{67.2}     \\
        \bottomrule
        \end{tabular} 
        }
\caption{Quantitative results concerning the domain adaptation approach and the usage of inter-coder connection. The bolded figure defines the best performance among the results.}
\label{tbl:condition_and_connection}
\vspace{-2ex}
\end{table}

\begin{table*}[t]
\resizebox{\textwidth}{!}{%
\centering
\begin{tabular}{c|c|ccccccccccccccccccc}
    \toprule
    Method & mIoU & Road & S.Walk & Build. & Wall & Fence & Pole & T. Light & T. Sign & Veget & Terrain & Sky & Person & Rider & Car & Truck & Bus & Train & M.Bike & Bike \\ \midrule
    \midrule
    \multicolumn{21}{c}{GTA5 $\rightarrow$ Cityscapes} \\
    \midrule
    \midrule
    AdaptSeg\cite{tsai2018learning}& 41.4 & 86.5 & 25.9 & 79.8 & 22.1 & 20.0 & 23.6 & 33.1 & 21.8 & 81.8 & 25.9 & 75.9 & 57.3 & 26.2 & 76.3 & 29.8 & 32.1 & 7.2 & 29.5 & 32.5 \\
    CBST\cite{zou2018unsupervised}& 45.9 & 91.8 & 53.5 & 80.5 & 32.7 & 21.0 & 34.0 & 28.9 & 20.4 & 83.9 & 34.2 & 80.9 & 53.1 & 24.0 & 82.7 & 30.3 & 35.9 & 16.0 & 25.9 & 42.8  \\
    DACS\cite{tranheden2021dacs} & 52.1  & 89.9 & 39.7 & 87.9 & 30.7 & 39.5 & 38.5 & 46.4 & 52.8 & 88.0 & 44.0 & 88.8 & 67.2 & 35.8 & 84.5 & 45.7 & 50.2 & 0.0 & 27.3 & 34.0 \\
    CorDA\cite{wang2021domain}& 56.6 & 94.7 & 63.1 & 87.6 & 30.7 & 40.6 & 40.2 & 47.8 & 51.6 & 87.6 & 47.0 & 89.7 & 66.7 & 35.9 & 90.2 & 48.9 & 57.5 & 0.0 & 39.8 & 56.0 \\
    BAPA\cite{liu2021bapa}& 57.4 & 94.4 & 61.0 & 88.0 & 26.8 & 39.9 & 38.3 & 46.1 & 55.3 & 87.8 & 46.1 & 89.4 & 68.8 & 40.0 & 90.2 & 60.4 & 59.0 & 0.0 & 45.1 & 54.2 \\
    ProDA\cite{zhang2021prototypical} & 57.5& 87.8 & 56.0 & 79.7 & 46.3 & 44.8 & 45.6 & 53.5 & 53.5 & 88.6 & 45.2 & 82.1 & 70.7 & 39.2 & 88.8 & 45.5 & 59.4 & 1.0 & 48.9 & 56.4  \\
    DAFormer\cite{hoyer2021daformer}& 68.3 & 95.7 & 70.2 & 89.4 & 53.5 & 48.1 & 49.6 & 55.8 & 59.4 & 89.9 & 47.9 & 92.5 & 72.2 & 44.7 & 92.3 & 74.5 & 78.2 & 65.1 & 55.9 & 61.8  \\
    HRDA \cite{hoyer2022hrda} & 73.8 & 96.4 & \textbf{74.4} & \textbf{91.0} & 61.6 & 51.5 & \textbf{57.1} & 63.9 & 69.3 & \textbf{91.3} & 48.4 & 94.2 & \textbf{79.0} & 52.9 & 93.9 & 84.1 & \textbf{85.7} & \textbf{75.9} & 63.9 & \textbf{67.5} \\
    \midrule
    \textbf{Our method} & \textbf{74.4} & \textbf{97.6} & 74.3 & 90.9 & \textbf{62.3} & \textbf{52.3} & 57.0 & \textbf{64.9} & \textbf{72.5} & 91.1 & \textbf{51.3} & \textbf{94.5} & 78.6 & \textbf{53.2} & \textbf{94.5} & \textbf{84.9} & 85.4 & 75.1 & \textbf{65.4} & 66.7  \\
    \midrule
    \midrule
    \multicolumn{21}{c}{Synthia $\rightarrow$ Cityscapes} \\
    \midrule
    \midrule
    AdaptSeg\cite{tsai2018learning}& 37.2  & 79.2 & 37.2 & 78.8 & - & - & - & 9.9 & 10.5 & 78.2 & - & 80.5 & 53.5 & 19.6 & 67.0 & - & 29.5 & - & 21.6 & 31.3 \\
CBST\cite{zou2018unsupervised}& 42.6 & 68.0 & 29.9 & 76.3 & 10.8 & 1.4 & 33.9 & 22.8 & 29.5 & 77.6 & - & 78.3 & 60.6 & 28.3 & 81.6 & - & 23.5 & - & 18.8 & 39.8  \\
DACS\cite{tranheden2021dacs}& 48.3 & 80.6 & 25.1 & 81.9 & 21.5 & 2.9 & 37.2 & 22.7 & 24.0 & 83.7 & - & 90.8 & 67.5 & 38.3 & 82.9 & - & 38.9 & - & 28.5 & 47.6  \\
CorDA\cite{wang2021domain}& 55.0 & 93.3 & 61.6 & 85.3 & 19.6 & 5.1 & 37.8 & 36.6 & 42.8 & 84.9 & - & 90.4 & 69.7 & 41.8 & 85.6 & - & 38.4 & - & 32.6 & 53.9  \\ 
BAPA\cite{liu2021bapa}& 53.3 & 91.7 & 53.8 & 83.9 & 22.4 & 0.8 & 34.9 & 30.5 & 42.8 & \textbf{86.8} & - & 88.2 & 66.0 & 34.1 & 86.6 & - & 51.3 & - & 29.4 & 50.5 \\
ProDA\cite{zhang2021prototypical}& 55.5 & 87.8 &  45.7 & 84.6 & 37.1 &  0.6 &  44.0 & 54.6 & 37.0 &88.1 & - & 84.4 &  74.2 &24.3 & 88.2 & - & 51.1 & - & 40.5 & 45.6 \\
DAFormer\cite{hoyer2021daformer}& 60.9 & 84.5 & 40.7 & 88.4 & 41.5 & 6.5 & 50.0 & 55.0 & 54.6 & 86.0 & - & 89.8 & 73.2 & 48.2 & 87.2 & - & 53.2 & - & 53.9 & 61.7 \\
HRDA \cite{hoyer2022hrda}& 65.8 & 85.2 & 47.7 & 88.8 & \textbf{49.5} & 4.8 & 57.2 & 65.7 & \textbf{60.9} & 85.3 & - & 92.9 & 79.4 & 52.8 & \textbf{89.0} & - & 64.7 & - & 63.9 & 64.9  \\
\midrule
\textbf{Our method}& \textbf{67.2} & \textbf{91.3} & \textbf{48.5} & \textbf{89.9} & \textbf{49.5} & \textbf{8.5} & \textbf{58.6} & \textbf{67.4} & 60.2 & 86.2 & - & \textbf{93.4} & \textbf{79.8} & \textbf{54.2} & 88.9 & - & \textbf{66.7} & - & \textbf{65.5} & \textbf{65.7}  \\
    \bottomrule    
\end{tabular}
} 
\vspace{1em}
\caption{Comparison with existing DA methods for UDA. The bolded figures represent the best performances.}
\label{tab:uda_comp}
\vspace{-2ex}
\end{table*}

\section{Ablation study}
\subsection{Effectiveness of the inter-coder connection}
\label{effective_intercoder}
The inter-coder connection facilitates the transfer of a wide range of latent features from the encoder to the decoder, drawing inspiration from the skip connections used in UNet \cite{weng2021inet} and ResNet \cite{he2016deep}. By concatenating both low-level and high-level features from the encoder, our architecture enhances the richness of feature representations and improves pixel-level segmentation accuracy—constituting a key structural distinction from standard latent diffusion models (LDMs).

To assess its effectiveness, we compared two segmentation models: one incorporating the inter-coder connection and one without it. The baseline model, which lacks the connection, uses reduced hidden layer dimensions in the decoder due to the absence of additional encoder-derived features. As shown in Table~\ref{tbl:condition_and_connection}, the model with the inter-coder connection achieved 74.4 mIoU on the GTA5$\rightarrow$Cityscapes benchmark and 69.3 mIoU on the Synthia$\rightarrow$Cityscapes benchmark, whereas the model without the connection obtained only 69.3 mIoU and 58.3 mIoU, respectively. The qualitative results in Figure~\ref{fig:quant_results_ablation} further illustrate that the inter-coder connection enables more accurate segmentation, particularly in the delineation of object boundaries.

These findings confirm that the inter-coder connection plays a significant role in enhancing segmentation performance within our proposed architecture.

\subsection{Effectiveness of adversarial learning in $\epsilon_{\theta}$}
The UDA process for ICCLD consists of two phases. The first phase involves domain adaptation for the encoder $\mathcal{E}$ and decoder $\mathcal{D}$ using a segmentation-based approach. The second phase performs domain adaptation on the denoising network $\epsilon_{\theta}$. Domain adaptation via segmentation—i.e., generating synthetic images and labels for training—is a widely adopted strategy in semantic segmentation UDA \cite{vayyat2022cluda, hoyer2022hrda, liu2021bapa}. Therefore, our ablation study focuses on examining the performance impact of the second phase, which applies adversarial learning to $\epsilon_{\theta}$.

As shown in Table~\ref{tbl:condition_and_connection}, the proposed adversarial learning applied to $\epsilon_{\theta}$ consistently improves segmentation performance. Regardless of whether the inter-coder connection is used, ICCLD models that include the second adaptation phase outperform those trained with only the segmentation-based phase. Specifically, ICCLD with the second phase achieves mIoU improvements of 9.0 and 3.8 on the GTA5$\rightarrow$Cityscapes and Synthia$\rightarrow$Cityscapes tasks, respectively. When the inter-coder connection is included, these gains increase to 8.0 and 6.5 mIoU, respectively.

Qualitative results in Figure~\ref{fig:quant_results_ablation} further support the benefit of applying adversarial learning to $\epsilon_{\theta}$. Models trained with adversarial learning produce more accurate segmentation outputs, notably reducing false positives. These findings suggest that the second domain adaptation phase helps mitigate the domain gap, enabling better adaptation to data variability and resulting in more accurate and robust performance.

\begin{figure*}
\begin{center}
\includegraphics[width=0.9\linewidth]{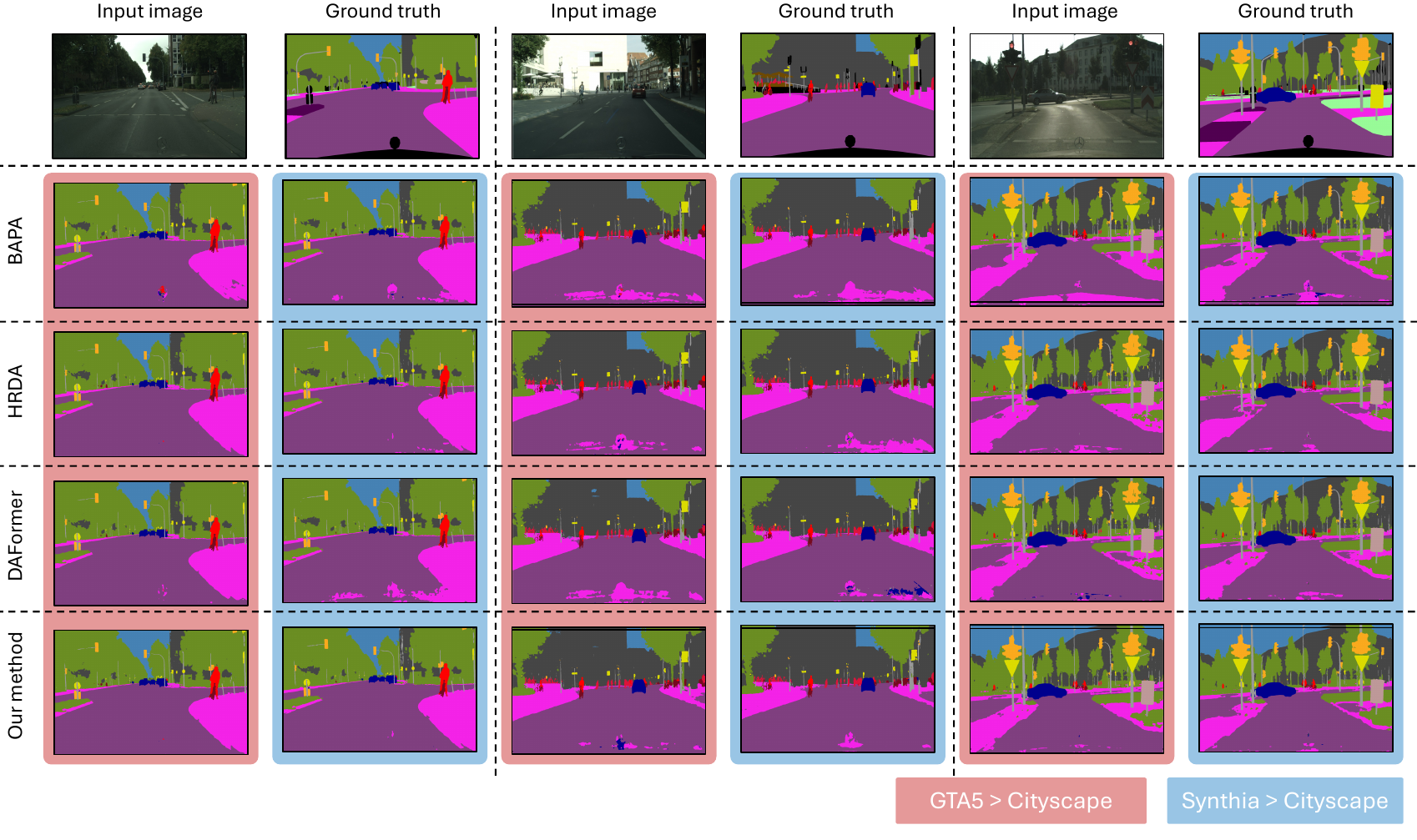}      
\end{center}
\caption{Qualitative comparison on the UDA performance using the proposed ICCLD with HRDA \cite{hoyer2022hrda}, DAFormer \cite{hoyer2021daformer}, and BAPA \cite{liu2021bapa}. HRDA and DAFormer achieve the 2$^{\text{nd}}$ and 3$^{\text{rd}}$ ranked performances on mIoU (See Table \ref{tab:uda_comp}).BAPA performs best for the Veget classes on the Synthia $\rightarrow$ Cityscapes UDA setting. The visualisation of the segmentation results shows that the proposed method produces more precise segmentation performance than the other methods.}
\label{fig:quant_results_comparison}
\vspace{-2ex}
\end{figure*}

\section{Comparison with existing UDA methods}
We begin by comparing the proposed approach with existing UDA methods \cite{tsai2018learning, zou2018unsupervised, tranheden2021dacs, wang2021domain, liu2021bapa, zhang2021prototypical, hoyer2021daformer, hoyer2022hrda}. As shown in Table~\ref{tab:uda_comp}, our method outperforms current SOTA techniques by a margin of +0.6 mIoU on the GTA5$\rightarrow$Cityscapes benchmark and +1.4 mIoU on the Synthia$\rightarrow$Cityscapes benchmark. Class-wise improvements are observed in 11 out of 19 classes for GTA5$\rightarrow$Cityscapes—particularly in challenging categories such as Wall, Rider, and Fence—and in 12 out of 16 classes for Synthia$\rightarrow$Cityscapes, as further quantified in Table~\ref{tab:uda_comp}.

Our method achieves 74.4 mIoU on the GTA5$\rightarrow$Cityscapes UDA setting and 67.2 mIoU on the Synthia$\rightarrow$Cityscapes setting, outperforming all other compared approaches. However, HRDA \cite{hoyer2022hrda} surpasses our method in a few specific classes in the GTA5$\rightarrow$Cityscapes benchmark, such as Sidewalk, Building, Pole, Vegetation, Person, Bus, Train, and Bike. Similarly, in the Synthia$\rightarrow$Cityscapes setting, BAPA \cite{liu2021bapa}, and HRDA show superior performance in a subset of categories. For instance, BAPA achieves 86.8 IoU for the Vegetation class, while HRDA achieves 60.9 IoU and 89.0 IoU for Traffic Sign and Car, respectively. Nevertheless, the performance differences for these individual classes are generally small—often below 0.2—indicating that our method maintains highly competitive results across the board.

In addition to quantitative comparisons, we provide qualitative analyses. Figure~\ref{fig:quant_results_comparison} presents segmentation results from our model alongside those of other high-performing methods. We include HRDA \cite{hoyer2022hrda} and DAFormer \cite{hoyer2021daformer}, which achieve the second- and third-highest mIoU scores, respectively. We also include visual results from BAPA \cite{liu2021bapa}, which attains the best performance for the Vegetation class in the Synthia$\rightarrow$Cityscapes benchmark (86.8 IoU; see Table~\ref{tab:uda_comp}). The visualisations indicate that our method produces cleaner and less noisy segmentation masks across both UDA settings. This suggests that it learns a more generalised and transferable representation from the source domain.

\section{Conclusion}
\label{sec:con}
In this paper, we introduce the Conditional and Inter-coder Connected Latent Diffusion (CICLD) model and adversarial learning for unsupervised domain-adaptive semantic segmentation.  CICLD is a variation of latent diffusion models, and it has inter-coder connections which bridge information obtained by an encoder to a decoder. The inter-coder connection helps to derive by delivering various abstracted features extracted from each layer of the encoder directly so that it improves the precision of segmentation results for complicated or small objects. In training UNet for the latent diffusion, we apply adversarial learning to reduce domain gaps by aligning the probabilistic character of the predicted noise distribution. Experimental results on GTA5$\rightarrow$Cityscapes and Synthia$\rightarrow$Cityscapes benchmarks show that CICLD achieves mIoU scores of 74.4 and 67.2, respectively, outperforming current SOTA methods.  

However, because our decoder now receives additional feature inputs from the encoder (via the inter-coder connection), its dimensionality increases, thereby raising computational overhead. In addition, the iterative denoising steps also prolong inference. In future work, we aim to optimize the architecture and the diffusion process to reduce this computational cost.

{
    \small
    \bibliographystyle{ieeenat_fullname}
    \bibliography{main}
}


\end{document}